\documentclass[preprint,times,authoryear,12pt]{els-workingpaper}
\usepackage[]{graphicx}\usepackage[]{color}
\makeatletter
\def\maxwidth{ %
  \ifdim\Gin@nat@width>\linewidth
    \linewidth
  \else
    \Gin@nat@width
  \fi
}
\makeatother

\definecolor{fgcolor}{rgb}{0.345, 0.345, 0.345}

\usepackage{framed}
\makeatletter
 {\par\unskip\endMakeFramed%
 \at@end@of@kframe}
\makeatother

\definecolor{shadecolor}{rgb}{.97, .97, .97}
\definecolor{messagecolor}{rgb}{0, 0, 0}
\definecolor{warningcolor}{rgb}{1, 0, 1}
\definecolor{errorcolor}{rgb}{1, 0, 0}
\newenvironment{knitrout}{}{} 

\usepackage{alltt}

\usepackage{amssymb,amsmath}

\newcommand{\stirlingsubset}[2]{\ensuremath{\genfrac{\lbrace}{\rbrace}{0pt}{}{#1}{#2}}}

\providecommand{\e}[1]{\ensuremath{\times 10^{#1}}}

\usepackage{graphicx}
\usepackage{xspace}
\usepackage{bm}
\usepackage{longtable}
\usepackage{hyphenat}
\usepackage{url}

\journal{1, submitted to Arxiv.org}
\msdate{January 3, 2013}
\IfFileExists{upquote.sty}{\usepackage{upquote}}{}
\begin{document}

\begin{frontmatter}

\title{Combinatorial Structure of the Deterministic Seriation Method with Multiple Subset Solutions}

\author{Mark E. Madsen}
\address{Department of Anthropology, Box 353100, University of Washington, Seattle WA, 98195 USA}
\ead{mark@madsenlab.org}
\ead[url]{http://notebook.madsenlab.org}

\author{Carl P. Lipo}
\address{Department of Anthropology and IIRMES, 1250 Bellflower Blvd, California State University at Long Beach, Long Beach CA, 90840 USA}
\ead{Carl.Lipo@csulb.edu}
\ead[url]{http://lipolab.org}

\begin{abstract}
Seriation methods order a set of descriptions given some criterion (e.g., unimodality or minimum distance between similarity scores).  Seriation is thus inherently a problem of finding the optimal solution among a set of permutations of objects.  In this short technical note, we review the combinatorial structure of the classical seriation problem, which seeks a single solution out of a set of objects.  We then extend those results to the iterative frequency seriation approach introduced by \citet{Lipo1997}, which finds optimal subsets of objects which each satisfy the unimodality criterion within each subset.  The number of possible solutions across multiple solution subsets is larger than $n!$, which underscores the need to find new algorithms and heuristics to assist in the deterministic frequency seriation problem. 
\end{abstract}


\end{frontmatter}

\section{Single Seriation Combinatorics}
\label{sec:single-seriation}



Seriation, whether employing class frequencies or simple occurrence to order assemblages, yields solutions which are permutations of the set of assemblages.  Because we cannot determine the ``polarity'' of a seriation solution---which ends represent early and late---from the class data alone, a unique seriation solution is thus formally a pair of mirror-image permutations:
\begin{equation}
\{a,d,b,c,e\} \equiv \{e,c,b,d,a\}
\end{equation}

This means that a set of $n$ assemblages can yield $n! / 2$ distinct solutions, regardless of whether solutions are composed of ordered similarity matrices or``Fordian'' frequency curves.  With small numbers of assemblages, enumeration and testing of all possible solutions is easy, even without parallel testing across many processors.  The ability to test solutions by enumeration quickly breaks down with only a modest number of assemblages.  Table \ref{tab:ss-stats} gives the number of unique solutions for selected problem sizes between 4 and 100 assemblages, and estimates of processing time to enumerate and test all solutions, assuming a cluster of 64 cores, and 0.005 seconds per solution test.\footnote{These assumptions concerning per-trial processing time and parallelism are arbitrary but within reach of social scientists given Amazon's EC2 cloud computing infrastructure, without requiring formal ``supercomputer'' access.  Modification by a factor of 10 has little effect on the results, perhaps shifting feasibility upward slightly before combinatorial explosion occurs.}  With 10 assemblages, we can test all solutions quickly enough that even a serial algorithm on a single core will be adequate to find the global best solution in a matter of hours, with parallelism improving this to real time responses.  

\begin{table}[ht]
\centering
\begin{tabular}{|c|r|r|r|}
  \hline
N & Seriation Solutions & Seconds & Years \\ 
  \hline
  4 &  12 & 0.00094 & 3e-11 \\ 
    6 & 3.6e+02 & 0.028 & 8.9e-10 \\ 
    8 & 2e+04 & 1.6 & 5e-08 \\ 
   10 & 1.8e+06 & 1.4e+02 & 4.5e-06 \\ 
   12 & 2.4e+08 & 1.9e+04 & 0.00059 \\ 
   13 & 3.1e+09 & 2.4e+05 & 0.0077 \\ 
   14 & 4.4e+10 & 3.4e+06 & 0.11 \\ 
   15 & 6.5e+11 & 5.1e+07 & 1.6 \\ 
   16 & 1e+13 & 8.2e+08 &  26 \\ 
   20 & 1.2e+18 & 9.5e+13 & 3e+06 \\ 
   40 & 4.1e+47 & 3.2e+43 & 1e+36 \\ 
   60 & 4.2e+81 & 3.3e+77 & 1e+70 \\ 
   80 & 3.6e+118 & 2.8e+114 & 8.9e+106 \\ 
  100 & 4.7e+157 & 3.6e+153 & 1.2e+146 \\ 
   \hline
\end{tabular}
\caption{Number of unique seriation solutions and parallel processing time for sets of assemblages $4 < n < 100$, testing solutions across 64 cores, assuming 5ms per trial} 
\label{tab:ss-stats}
\end{table}


A typical characteristic of many combinatorial algorithms is that small changes in problem size can have massive changes in processing time.  13 assemblages will turn out to be the practical limit for direct enumeration, even given parallel processing with circa-2012 technology, with total processing time of nearly 3 days running 64 cores at full capacity.\footnote{Realistically, almost nobody would contemplate doing this, given the expense of the computing time relative to the value of guaranteeing the optimal solution, but the hypothetical example demonstrates that such solutions are \emph{feasible}.}  Problems involving 14 and 15 assemblages reach the point where large clusters require more than a month and 19 months respectively, to solve.  Beyond 15 assemblages, a ``combinatorial explosion'' sets in, with 20 assemblages requiring more than 3 million years, before solution times quickly exceed the lifetime of the universe.  

In short, top-down enumerative methods are feasible for small sets of assemblages, and given widespread availability of multiple core computers, seriation packages should employ enumeration for small problems, or to build and test smaller parts of larger seriation solutions.  

\section{Deterministic Seriation with Multiple Solution Groups}
\label{sec:seriation-groups}



\begin{figure}
	\includegraphics[scale=0.75]{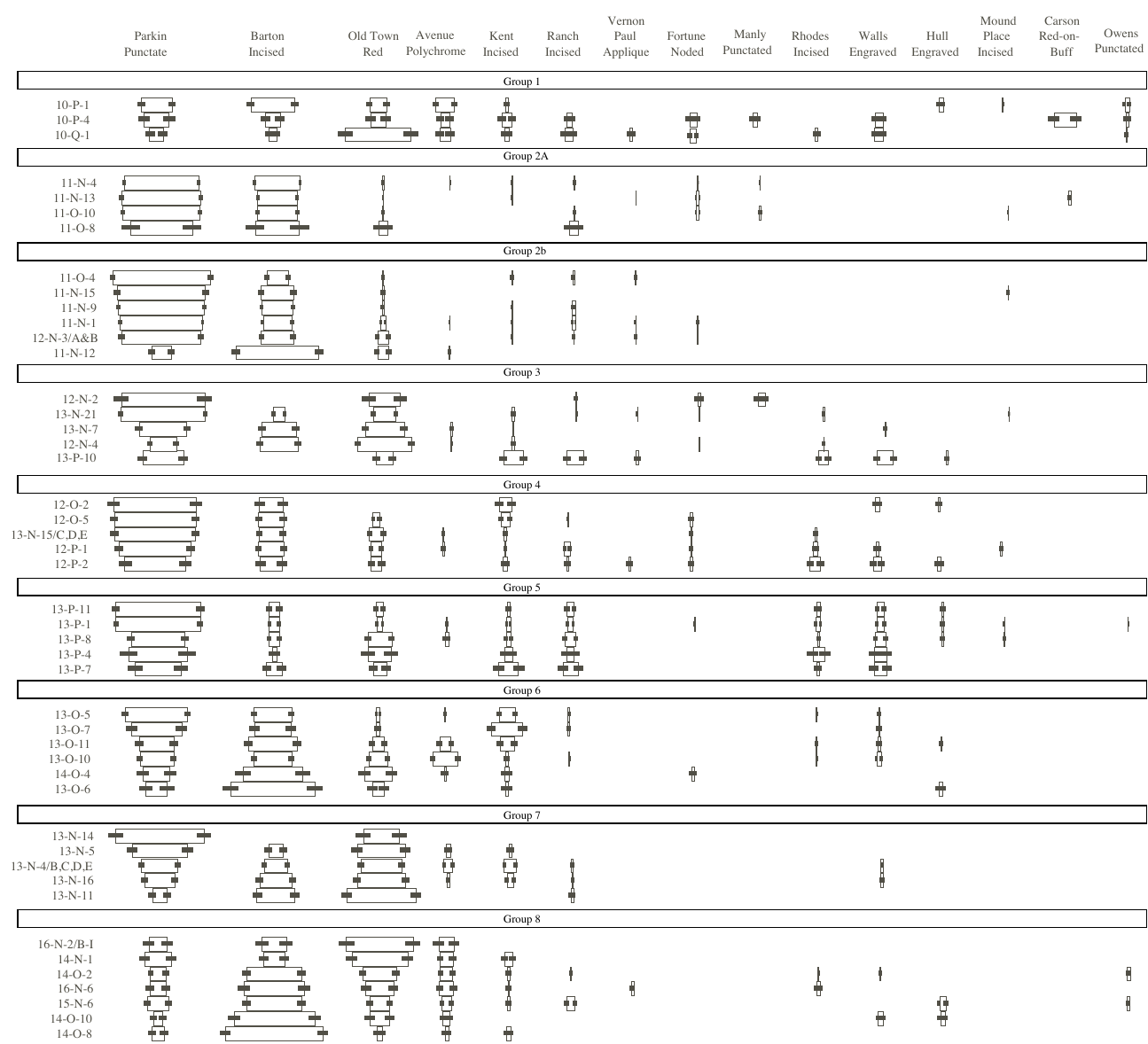}
	\caption{Example of a deterministic frequency seriation with assemblages partitioned into multiple subsets or solution groups.  From \citet{Lipo2001b}, Figure 4.4.}
	\label{fig:mult-seriation-groups}
\end{figure}

In an earlier paper \citep{Lipo1997}, we introduced an iterative method for finding deterministic solutions to the frequency seriation problem by partitioning assemblages into subsets, each of which meets the unimodal ordering principle, within tolerance limits governed by sample size.  \citet{Lipo2001b} extended and refined the method in his dissertation research.  Our initial work on the method employed a combination of automated calculations (e.g., bootstrap significance tests for pairwise orderings), and manual sorting of assemblages into groups and specific positions (using an Excel macro package available at \url{http://lipolab.org/seriation.html}).  Figure \ref{fig:mult-seriation-groups} is an example of seriation with multiple solution groups, from Lipo's dissertation research in the Lower Mississippi Valley.  

Our initial work suggests assemblages seriate together into groups reflecting variation in the intensity of cultural transmission among assemblages, over their duration of accumulation. In most cases, solution groups tend to be spatiotemporally compact, and form clusters when mapped on the landscape, although long-distance connections between past communities can also yield patterns which are more complex and less cohensive when mapped.  Madsen's dissertation research is aimed at tying the properties seriation solution groups to their causes in regional patterns of interaction and the dynamics of specific cultural transmission models.  

\begin{table}[ht]
\centering
\begin{tabular}{|c|r|r|r|}
  \hline
\# of Solution Groups (m) & 20 & 40 & 60 \\ 
  \hline
  3 & 5.8e+08 & 2e+18 & 7.1e+27 \\ 
    4 & 4.5e+10 & 5e+22 & 5.5e+34 \\ 
    6 & 4.3e+12 & 1.8e+28 & 6.8e+43 \\ 
    8 & 1.5e+13 & 3.2e+31 & 3.8e+49 \\ 
   10 & 5.9e+12 & 2.4e+33 & 2.7e+53 \\ 
   15 &  & 2.9e+34 & 2.2e+58 \\ 
   20 &  & 1.6e+32 & 1.7e+59 \\ 
   25 &  &  & 3.7e+57 \\ 
   30 &  &  & 9.6e+53 \\ 
   \hline
\end{tabular}
\caption{Number of ways to form m subsets (seriation solutions) from 20, 40, and 60 assemblages} 
\label{tab:subsets}
\end{table}


In this section, the goal is to understand the complexity of the multiple seriation groups problem, constructing reasonable upper bounds for a given problem size, even if some problems encountered in real analyses do not approach the worst case.  From a combinatorial standpoint, seriation with multiple solution groups has the following structure.  
We begin with $n$ assemblages in total, and seek a solution or solutions whereby we end up with $m$ solution groups, where $m < n$.  Each solution must have at least one assemblage, and in practice will often have 3 or more (singletons may indicate assemblages which simply do not ``fit'' with anything else in the data set).  The number of ways that $n$ objects can be partitioned into $m$ non-empty subsets (or solution groups) is given by the Stirling numbers of the second kind, which are given by the recursion equation:
\begin{equation}
\stirlingsubset{n}{m} = m \stirlingsubset{n-1}{m} + \stirlingsubset{n-1}{m-1}
\end{equation}
Table \ref{tab:subsets} gives the number of ways to form a specific number of subsets (or seriation solution groups) from sets of assemblages ranging from 20 to 60.  Each column runs from 3 solution groups to half of the number of assemblages, since the number of possible subsets is maximized just before $n/2$ and declines thereafter (Figure \ref{fig:subsets-graph}).  

\begin{knitrout}
\definecolor{shadecolor}{rgb}{0.969, 0.969, 0.969}\color{fgcolor}\begin{figure}[h!]
\includegraphics[width=3.5in]{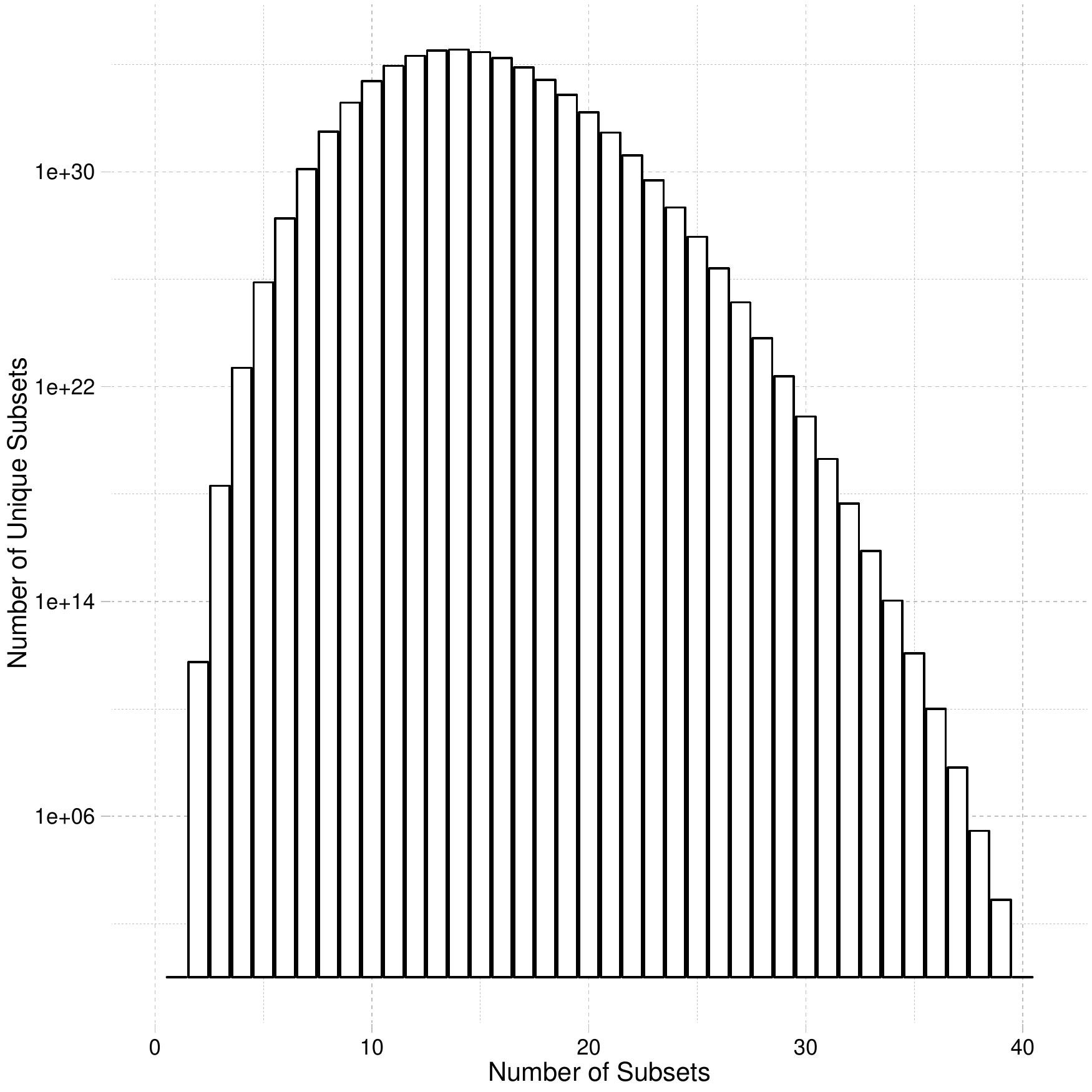} \caption[Number of Unique Solution Sets for 40 Assemblages When Partitioned Into ]{Number of Unique Solution Sets for 40 Assemblages When Partitioned Into $m$ Solution Groups\label{fig:subsets-graph}}
\end{figure}

\end{knitrout}

We can immediately see that there are an enormous number of possible subsets for any assemblage size.  There are fewer subsets, of course, than complete permutations of the set of assemblages since subsets are unordered (i.e., $\stirlingsubset{n}{m} < n! \,\mathrm{for\,all }\,m$).  However, in the multiple seriation group problem, the problem size is larger than the corresponding Stirling number because we do not know in advance how many groups (subsets) a set of assemblages will seriate into.  Thus, the total number of unique subsets which might contain the optimal solution is the total of the number of subsets, across all subset sizes:
\begin{equation}
\sum_{i=1}^n \stirlingsubset{n}{i}
\label{eq:sum-stirling}
\end{equation}
This result is still smaller than the total permutations for a set of $n$ assemblages.  For example, given 40 assemblages, $n! = 8.159\e{47}$, whereas the total from Equation \ref{eq:sum-stirling} for 40 assemblages is $1.575\e{35}$.  

Another factor to consider is that each of these unique subsets resulting from a partition of $n$ assemblages into seriation groups is still unordered.  For example, if we partition 10 assemblages into 3 solution groups, there are 9330 unique ways of assigning the 10 assemblages to the 3 solution groups.  Each group within a partition will have $n_i$ members, where $\sum n_i = n$.   The number of unique seriations for each of the 3 solution groups is $n_i ! / 2$, but we cannot assume that solution groups will have a balanced or equal number of assemblages (as Figure \ref{fig:mult-seriation-groups} does).  Partitions such as:
\begin{equation*}
\{1,2,3,4,5,6\} \{7,8\} \{9,10\}  
\end{equation*}
are common in seriating real assemblages \citep{Lipo2001b}.   





Since the factorial function grows so quickly, the computational cost of determining the correct permutation within a given seriation solution group is controlled by the size of the largest subset, especially if the other subsets are relatively small, as in the previous example.  At worst, for a solution set with $m$ solution groups, $m-1$ solution groups will contain 1 assemblage each, and the last solution group will consist of the remaining $n-m-1$ assemblages.  This means, of course, that the worst case would involve consideration of on the order of $(n-m-1)!$ permutations within each solution group, for each of the subsets given by Equation \ref{eq:sum-stirling}.  This yields:
\begin{equation}
\sum_{m=1}^n \stirlingsubset{n}{m} (n-m-1)!
\end{equation}
Table \ref{tab:total-mult} gives the total number of possible solutions for assemblages ranging from 4 to 100, where solutions may fall into multiple seriation groups of any size.  

\begin{table}[ht]
\centering
\begin{tabular}{|c|r|r|r|}
  \hline
N & Total Solutions & Seconds & Years \\ 
  \hline
  4 &  15 & 0.0012 & 3.7e-11 \\ 
    6 & 4.7e+02 & 0.037 & 1.2e-09 \\ 
    8 & 5.2e+04 &   4 & 1.3e-07 \\ 
   10 & 1.5e+07 & 1.1e+03 & 3.6e-05 \\ 
   12 & 8.5e+09 & 6.6e+05 & 0.021 \\ 
   13 & 2.6e+11 & 2e+07 & 0.64 \\ 
   14 & 8.9e+12 & 7e+08 &  22 \\ 
   15 & 3.5e+14 & 2.8e+10 & 8.7e+02 \\ 
   16 & 1.6e+16 & 1.2e+12 & 3.9e+04 \\ 
   20 & 1.7e+23 & 1.3e+19 & 4.2e+11 \\ 
   40 & 9e+65 & 7e+61 & 2.2e+54 \\ 
   60 & 5.1e+116 & 4e+112 & 1.3e+105 \\ 
   80 & 5.1e+172 & 4e+168 & 1.3e+161 \\ 
  100 & 4.4e+232 & 3.4e+228 & 1.1e+221 \\ 
   \hline
\end{tabular}
\caption{Number of total solutions with multiple seriation groups and processing time for sets of assemblages 4 < n < 100, testing solutions across 64 cores} 
\label{tab:total-mult}
\end{table}


\section{Discussion}
\label{sec:conclusions}

Clearly, in the worst case, the combinatorial complexity of the multiple seriation groups problem is much worse than even the straight factorial case involved in single solution permutations.  The feasibility of parallelized enumerative methods still explodes after 13 assemblages, but much more steeply.  The goal of a new algorithm for deterministic multiple group seriations is, therefore, to employ heuristics to drastically reduce the size of the solution space.  Vast amounts of the solution space involve partial orders which violate unimodality, but of course we cannot easily identify those regions of solution space \emph{a priori} without testing possibilities.  But given small partial solutions which do meet the seriation model, we can easily test solutions which are ``adjacent'' to the partial solutions, suggesting that agglomerative heuristics may be the best approach to finding a computationally feasible method.

\section*{References Cited}

\bibliographystyle{model2-names}
\bibliography{seriation-combinatorics}

\begin{thebibliography}{2}
\expandafter\ifx\csname natexlab\endcsname\relax\def\natexlab#1{#1}\fi
\expandafter\ifx\csname url\endcsname\relax
  \def\url#1{\texttt{#1}}\fi
\expandafter\ifx\csname urlprefix\endcsname\relax\def\urlprefix{URL }\fi
\providecommand{\eprint}[2][]{\url{#2}}
\providecommand{\bibinfo}[2]{#2}
\ifx\xfnm\relax \def\xfnm[#1]{\unskip,\space#1}\fi
\bibitem[{Lipo et~al.(1997)Lipo, Madsen, Dunnell and Hunt}]{Lipo1997}
\bibinfo{author}{Lipo, C.}, \bibinfo{author}{Madsen, M.},
  \bibinfo{author}{Dunnell, R.}, \bibinfo{author}{Hunt, T.},
  \bibinfo{year}{1997}.
\newblock \bibinfo{title}{Population structure, cultural transmission, and
  frequency seriation}.
\newblock \bibinfo{journal}{Journal of Anthropological Archaeology}
  \bibinfo{volume}{16}, \bibinfo{pages}{33}.
\bibitem[{Lipo(2001)}]{Lipo2001b}
\bibinfo{author}{Lipo, C.P.}, \bibinfo{year}{2001}.
\newblock \bibinfo{title}{Science, Style and the Study of Community Structure:
  An Example from the Central Mississippi River Valley}.
\newblock \bibinfo{publisher}{British Archaeological Reports, International
  Series, no. 918}, \bibinfo{address}{Oxford}.

\end{thebibliography}

\end{document}